\documentclass[runningheads]{llncs}
\usepackage[T1]{fontenc}
\usepackage{pgfplots}
\usepackage{graphicx}
\usepackage{cite}
\usepackage[hidelinks]{hyperref}
 \usepackage{hyperref}
\usepackage{multirow}
\usepackage{appendix}
\usepackage[linesnumbered,ruled,vlined]{algorithm2e}

\usepackage{amsmath,amsfonts,amssymb,amscd,xspace}
\SetKwInput{KwInput}{Input}                
\SetKwInput{KwOutput}{Output}  
\begin{document}
\title{REFINE on Scarce Data: Retrieval Enhancement through Fine-Tuning via Model Fusion of Embedding Models}
\titlerunning{REFINE on Scarce Data}
%
\author{Ambuje Gupta\inst{1}\orcidID{0009-0000-4230-1440} \and
Mrinal Rawat\inst{1}\orcidID{0000-0003-4900-0733} \and
Andreas Stolcke\inst{1}\orcidID{0000-0002-9925-905X} \and
Roberto Pieraccini\inst{1}\orcidID{0009-0009-2608-6184}}

%
%
\authorrunning{Gupta et al.}
\institute{Uniphore, Palo Alto, CA 94304, USA \\
\email{mrinal.rawat@uniphore.com}\\
 }
\maketitle              
\begin{abstract}
Retrieval augmented generation (RAG) pipelines are commonly used in tasks such as question-answering (QA), relying on retrieving relevant documents from a vector store computed using a pretrained embedding model. However, if the retrieved context is inaccurate, the answers generated using the large language model (LLM)  may contain errors or hallucinations. Although pretrained embedding models have advanced, adapting them to new domains remains challenging. Fine-tuning is a potential solution, but industry settings often lack the necessary fine-tuning data. To address these challenges, we propose REFINE, a novel technique that generates synthetic data from available documents and then uses a model fusion approach to fine-tune embeddings for improved retrieval performance in new domains, while preserving out-of-domain capability. 
We conducted experiments on the two public datasets: SQUAD and RAG-12000 and a proprietary TOURISM dataset. Results demonstrate that even the standard fine-tuning with the proposed data augmentation technique outperforms the vanilla pretrained model. Furthermore, when combined with model fusion, the proposed approach achieves superior performance, with a \textbf{5.76\%} improvement in recall on the TOURISM dataset, and \textbf{6.58 \%} and \textbf{0.32\%} enhancement on SQUAD and RAG-12000 respectively.

\end{abstract}

\section{Introduction}
Question Answering (QA) systems for enterprise data have been a longstanding research focus \cite{article}. Many efforts have centered on using BERT-based models that predict the start and end points of answer spans from the given context \cite{joshi2020spanbert}. While initially promising, in practical business settings, there is an increasing need for abstractive QA that can provide more nuanced responses beyond extracting spans. 

The rise of large language models (LLMs), such as ChatGPT \cite{ouyang2022training} and Llama2 \cite{touvron2023llama}, has advanced abstractive QA capabilities substantially by enabling more sophisticated response generation. However, a key challenge with LLMs is their lack of up-to-date, real-world knowledge, and tendency to hallucinate with confidence \cite{xu2024hallucination}. To alleviate this issue, retrieval-augmented generation (RAG) \cite{10.5555/3495724.3496517} has emerged as a promising technique that augments LLMs with relevant contextual information retrieved from databases. However, RAG's effectiveness depends on providing accurate context---if the retrieved context is flawed, errors can propagate into the generated responses. The retrieval process in RAG typically involves ingesting documents, segmenting them, and representing them using embedding models such as BGE \cite{bge_m3}. These representations are then stored in vector databases. At run-time, the query embedding is computed and matched against the pre-computed document embeddings to retrieve the most relevant matches.

A key challenge is that general pretrained embedding models may not perform well for domain-specific enterprise data. Moreover, these models may fail to provide precise results for retrieval use cases as their training objective did not focus specifically on matching queries to relevant documents. 
One potential solution is to fine-tune embeddings using domain-specific dataset, but obtaining labeled data is difficult in many application scenarios. Although unlabeled documents are usually available, they can still be scarce in practical scenarios, complicating the situation further.  Some approaches \cite{gururangan-etal-2020-dont,huang-etal-2023-adasent}  have therefore employed continued pretraining on unlabeled documents. However, this method requires a substantial amount of data, which may not be available.

To address these problems, we propose REFINE---a novel approach to enhance retrieval performance using unlabeled documents, particularly when data is scarce. Our contributions can be summarized as follows:
\begin{itemize}

\item  A novel approach to generate contrastive training datasets for fine-tuning embedding models suited for retrieval use-cases, using only available unlabeled documents, by leveraging an LLM.

\item  Second, we demonstrate how the generated dataset can enhance the embedding model through standard fine-tuning methods, without any additional techniques.

\item  Third, we introduce a model fusion technique during fine-tuning that incorporates both pretrained representations and the new data-specific learning, boosting performance further while retaining general capabilities for out-of-domain datasets.
\end{itemize}

\section{Related Work}
Document retrieval systems have revolutionized information access, allowing users to find relevant content from large databases quickly. These systems are particularly critical in question-answering (QA) tasks, where relevant documents provide the necessary context for generating accurate answers \cite{voorhees-tice-2000-trec}. The emergence of large language models (LLMs) has increased the importance of document retrieval, especially in RAG applications. The LLM context is augmented with the right documents so that it generates the correct answers without hallucinations \cite{10.5555/3495724.3496517}. The effectiveness of such retrieval systems often depends upon the quality of the document embeddings that are used to represent and retrieve the relevant information. Initially, systems such as BM-25, which employ sparse retrievers, were common \cite{INR-019}. Although effective when exact word matches are present, they struggle with semantic understanding. To address this, the use of dense retrievers became prevalent. These retrievers, which are essentially embedding models, provide semantic text representations. DPR \cite{karpukhin-etal-2020-dense} demonstrates the practicality of using dense representations for retrieval with a dual encoder framework trained on a modest set of questions and passages.

A major breakthrough in embedding models occurred with the introduction of BERT-based models \cite{devlin-etal-2019-bert}, which provide contextual representations through Masked Language Modeling (MLM), a self-supervised learning technique. Subsequently, considerable effort was made to create generalized embeddings, such as Sentence Transformers e.g. MPNet \cite{reimers-2019-sentence-bert}, OpenAI embeddings, and BGE embeddings \cite{xiao2024cpack}. These embeddings trained on a vast corpus using contrastive learning have demonstrated significant improvements. However, in many scenarios, they could still fail to correctly represent the text if the domain is very specialized. In these cases, continual pretraining and fine-tuning can be very helpful. Usually, if the domain-specific dataset is large, the model pretraining is first continued in a self-supervised fashion \cite{gururangan-etal-2020-dont}, which does not require labeled data. Then, the model is fine-tuned on the target task to further enhance performance for that specific domain and task. However, this fine-tuning requires labeled data, preventing its use in many data-starved application domains. Synthetic data generation has been proposed as a solution to the data scarcity problem. By generating artificial samples that mimic the characteristics of a target domain, the natural training set is augmented, allowing adaptation to new domains without extensive new annotation \cite{wu2024llmaugmented, thakur-2020-AugSBERT}. 

Furthermore, while fine-tuned models excel on the domains they are trained on, they can lead to a significant drop in performance on more general tasks, a phenomenon known as catastrophic forgetting \cite{goodfellow2015empirical, recadam}. To counteract this, various strategies have been proposed, including replay-based methods \cite{rolnick2019experience} and regularization-based methods \cite{rebuffi2017icarl}. More recently, the concept of model merging has been shown effective to harness the strengths of different models. This technique involves combining models that have been trained on different datasets or that employ distinct architectures \cite{cocktail}. However, this merging typically occurs after fine-tuning and without continued training. Our proposed REFINE method, by contrast, incorporates model merging during the training phase. This strategy allows for more effective representation learning as the model continues using knowledge from the pretrained model while adapting to the new domain.

\begin{figure*}[ht]
\centering
     \includegraphics[trim={0 1cm 0 1.1cm},clip,scale=0.37]{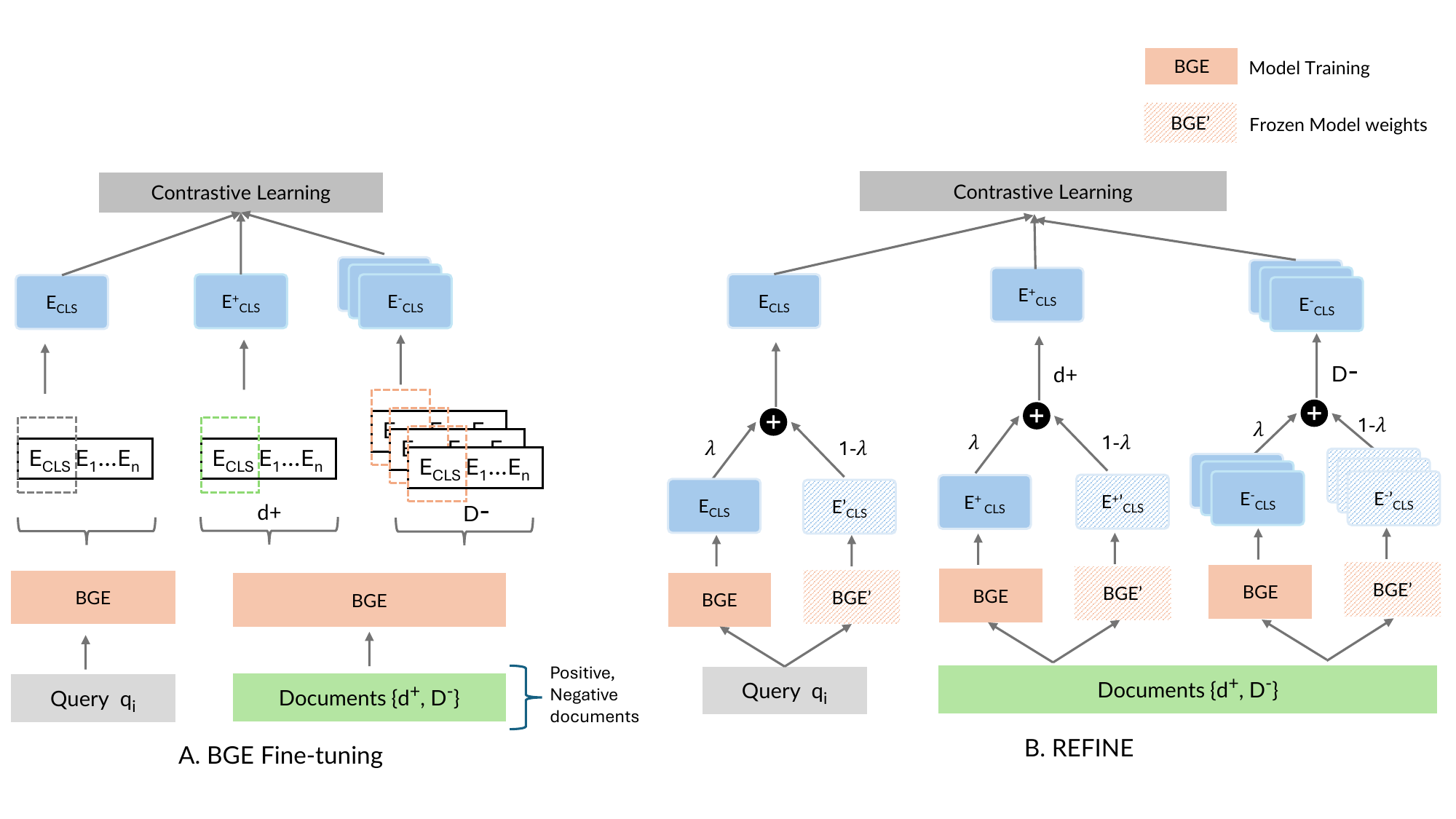}
      \caption{(A) standard BGE process followed for fine-tuning. (B) training process for REFINE.}
       \label{refine_training}
\end{figure*}

\section{Approach}
In this work, we introduce a data augmentation technique employed to fine-tune the embedding model. This technique aims to enrich the representations and enhance the retrieval performance. Additionally, we propose a model fusion technique to further improve the fine-tuning process. In the subsequent subsections, we will elaborate on these concepts in detail.

\subsection{Data Augmentation for Fine-tuning}
\label{subsection:data_aug}

LLMs have been utilized for data augmentation or synthetic data creation, having shown good out-of-the-box performance. This process involves automating the laborious and time-consuming task of creating and annotating data by humans. With LLMs, we can generate synthetic data rapidly, and in some cases, even outperform human text annotation quality \cite{Gilardi_2023}. With the advent of larger and newer LLMs, the quality of LLM-generated annotations and data can be expected to improve further.

We utilize LLM to generate high-quality synthetic data that is used to fine-tune the embedding model, BGE in our case. We start with a set of given unlabeled documents $D$, and for each document, we prompt the LLM to generate a set of derived queries $Q = \{q_1, q_2, \ldots, q_k\}$. The prompt we used for generating queries is:
\begin{quote}
\texttt{You are a query generator bot. Generate 10 distinct queries from the document $D$}
\end{quote}

In our experiments, we set $k$ to 10. Now, for each generated query $q_i \in Q$, since we know from which document the question was generated, its corresponding source document becomes the positive sample passage $d^{+}_i$. For contrastive fine-tuning of the embedding model, we also require negative sample passages, so that the model learns to discriminate between positive and negative matches between queries and passages.

\begin{algorithm}[th]
\DontPrintSemicolon
\KwInput{Documents $D$, pretrained BGE model $M$}

\SetKwFunction{SearchF}{Evaluation}
\SetKwProg{Fn}{Function}{:}{}
$D_A \gets \emptyset$; \tcp*{Augmented Dataset}
$DB \gets create\_vectordb(D, M)$ \\
\ForEach{$d^+_i \in D$}{
    $Q \gets generate\_queries(d^+_i,k=10)$ \tcp*{Generates 10 queries using LLM from the document}
    \ForEach{$q_i \in Q$}{
        $R \gets retrieve\_docs(q_i, DB)$ \\
        $D^- \gets select\_neg\_docs(R)$ \tcp*{pick hard negative documents with similarity scores obtained from retriever between 0.5 and 0.7, excluding those in the top 5.}
        $D_A \gets D_A \cup  \{ (q_i, d^+_i, D^-) \}$ 
    }
}

\caption{Pseudo-code for data augmentation}
\vspace{-0.15cm}
\label{algo}
\end{algorithm}

We aim to find ``hard negative'' training samples to enhance the model's discriminative power. To that end, we first vectorize and ingest all documents to create a vector store, using a pretrained embedding model. Now, for each query $q_i \in Q$, we use the retriever to locate the top 50 most similar documents. Next, we filter the documents, selecting those that have similarity scores between 50\% and 70\% and are not in the top $k = 5$, to keep near misses while ruling out accidental matches. We found these hyperparameters empirically by running trials on a held-out 15\% of the training set.  In this way, we obtain $D^- = \{ d^{-}_1, d^{-}_2, \ldots, d^{-}_m \}$ as negative training samples (cf.\ Algorithm~\ref{algo}).

\subsection{Fine-tuning via Model Fusion}

Contrastive learning on diverse data has demonstrated good performance in similarity learning based on embedding models, as it learns to push negative samples away from the query, while pulling positive samples closer to it. Although refining the model through additional fine-tuning can improve its effectiveness within a particular domain \cite{xiao2024cpack}, it is important to note that this specialization might deteriorate performance outside of the fine-tuning domain.

In this work, we also utilize contrastive learning to fine-tune the model using a model fusion approach (Figure~\ref{refine_training}); specifically the \texttt{BAAI/bge-large-en-v1.5} model, is fine-tuned on a domain-specific dataset obtained using the technique discussed in Section \ref{subsection:data_aug}. We first initialize two models, $BGE$ and $BGE'$, with the pretrained BGE weights, freezing the weights of $BGE'$ while allowing only $BGE$ to undergo fine-tuning \cite{xiao2024cpack}. Given a query $q_i$, we pass it through both $BGE$ and $BGE'$ to obtain the representations of the \texttt{[CLS]} token, denoted as $E_{CLS}$ and $E'_{CLS}$, respectively. Next, we interpolate these two representations to create a final representation $E_{CLS}$:
\begin{equation} \label{eq:combine_embeddings}
\begin{aligned}
    \textit{E}_{CLS} = \lambda \cdot E_{CLS} +  (1-\lambda) \cdot E'_{CLS}
\end{aligned}
\end{equation}
where $\lambda$ is a hyperparameter. Increasing its value will place greater emphasis on the new domain.

Next, we apply the same technique to the positive passage $d^+_i$ and the negative passages $D^-_i$, yielding $E^+_{CLS}$ and \{$E^-_{CLS_1}$, $E^-_{CLS_2}$, \ldots, $E^-_{CLS_m}\}$, respectively. We then fine-tune the model using the objective

\begin{equation} \label{eq:cl}
{\max \sum \log \frac{e^{sim(E_{CLS}, E^+_{CLS})/\tau}}{e^{sim(E_{CLS}, E^+_{CLS})/\tau} + \sum_{i=1}^{m}{e^{sim(E_{CLS}, E^-_{CLS_i})/\tau}}}}
\end{equation}

where $\tau$ is a temperature hyperparameter and $sim()$ is the cosine similarity.

\section{Experiments}
\subsection{Datasets}

We conducted experiments on two public and one propriety datasets: 

\textbf{SQUAD\footnote{\href{https://huggingface.co/datasets/rajpurkar/squad}{rajpurkar/squad}}} is a question-answering dataset along with the provided context. To simulate limited dataset conditions, we utilized a subset comprising 300 randomly selected instances from the publicly available dataset. 

\textbf{RAG 12000\footnote{ \href{https://huggingface.co/datasets/neural-bridge/rag-dataset-12000}{RAG 12000 Dataset}}} is a collection of triples featuring \texttt{context}, \texttt{question}, and \texttt{answer} fields, designed for building models optimized for RAG. Similar to SQUAD dataset, we took a subset of 100 randomly selected instances from the dataset. We will refer to it as the RAG dataset for brevity.

\textbf{TOURISM} is a proprietary dataset focusing on the tourism domain. It contains both unstructured text and structured, tabular data.

To evaluate the retriever, we require both the question and the supporting document (evidence). With the SQUAD \& RAG dataset, this is straightforward due to the inherent mapping present within the raw dataset, allowing us to create a dataset of question-document (context) pairs for evaluation. However, for the TOURISM dataset, our annotation team manually curated a question from each document. For fine-tuning, we generated a dataset exclusively from documents and synthetic questions. To perform hyperparameter tuning, we set aside 15\% of the generated data as a validation set for each of the datasets. Table \ref{tab:stats} summarizes the sizes of these datasets.

\begin{table}[th!]
\caption{Statistics of the two datasets, with the number of documents and questions for evaluation, as well as training and validation (tuning) samples generated through our data augmentation method.}
\centering
\begin{tabular}{lccccc}
\hline
Dataset & \# Docs & \# Ques. & \# Train & \#Val & Avg Words (Docs)\\
\hline
SQUAD   & 300        & 300    & 2550   & 450  & 165       \\
RAG 12000   & 100        & 100    & 850   & 150  & 190       \\
TOURISM & 58         & 58     & 493      & 87 & 138  \\ \hline
\end{tabular}

\label{tab:stats}

\end{table}

\subsection{Baselines}
\label{baselines}
We present a direct comparison of our approach with the following baseline methods:

\begin{itemize}
\item \textbf{Vanilla BGE:} Here, we compute the metrics by simply utilizing the pretrained BGE\footnote{ \href{https://huggingface.co/BAAI/bge-large-en-v1.5}{BAAI/bge-large-en-v1.5}} embedding model on the two evaluation sets.

\item \textbf{Vanilla snowflake-arctic-embed-l:\footnote{ \href{https://huggingface.co/Snowflake/snowflake-arctic-embed-l}{Snowflake/snowflake-arctic-embed-l}}} We use an embedding model that attained state-of-the-art performance on the MTEB leaderboard \cite{merrick2024arcticembed}. 

\item \textbf{Vanilla text-embedding-3-large:} Metrics are computed using a state-of-the-art off-the-shelf text embedding model \cite{openai-textembed}.

\item \textbf{Fine-tuned BGE:} We fine-tune the BGE model according to the method outlined by \cite{xiao2024cpack} on the dataset generated synthetically with our approach, and subsequently compute the metrics on the evaluation data. Figure \ref{refine_training}(A) illustrates this technique.

\item \textbf{LM Cocktail:} We performed model merging after training \cite{cocktail}, combining vanilla BGE with the fine-tuned BGE model. We applied the same weighting as in our REFINE model, i.e., 0.35 for BGE and 0.65 for fine-tuned BGE.

\end{itemize}

\subsection{Evaluation Metrics}
\label{sec:eval_met}
We use the following commonly used metrics for evaluating the retriever:

\begin{itemize}
\item \textbf{Mean average precision (MAP)} measures the average precision across multiple queries. Higher values indicate better performance, rewarding systems that return relevant documents earlier in the ranked list.

\item \textbf{Normalized discounted cumulative gain (NDCG)} assesses ranking quality by considering relevance and position in the ranked list. Higher scores are given to highly relevant documents at the top.

\item \textbf{Mean reciprocal rank (MRR)} evaluates systems producing ranked lists by considering the rank of the first relevant document. Higher values indicate the relevant document is ranked higher.

\item \textbf{Recall} measures the fraction of relevant documents successfully retrieved by the system. Higher values indicate better performance in retrieving all relevant document
\end{itemize}

\subsection{Evaluation Setup}
To evaluate the retriever's performance, we initially ingest documents from each dataset into a FAISS vector store \cite{douze2024faiss}, separately. Then, for every query in the question set, we derived its representation using the model fine-tuned through our proposed method. Subsequently, we used this representation to fetch relevant documents from the vector store, utilizing the retriever from the LangChain library. For most experiments, and unless indicated otherwise, we fixed the top\_k parameter at 3 (retrieving the three most relevant documents). (The number of retrieved documents is commonly notated as ``@k'' following the metric, e.g., ``Recall@3.'').  We computed the evaluation metrics for all the baseline approaches described in Section \ref{baselines}. We want to highlight that we did not concentrate on hyperparameter tuning. However, to account for the variability in results due to different random initializations, we report the average performance across five experimental runs with different random seeds for the models that involved training.

\subsection{Training Details}
In this work, we leveraged the \texttt{BAAI/bge-large-en-v1.5} as the backbone model. To train both the BGE way and REFINE, identical hyperparameters were used. The learning rate was set to $10^{-5}$, and the batch size was adjusted according to GPU availability, with 4 gradient accumulation steps implemented for both approaches. Notably, in the case of REFINE, we froze the layers of $BGE'$ while allowing gradient updates for $BGE$. In the REFINE approach, a crucial parameter is the value of $\lambda$, which we set at 0.35, chosen empirically. This implies that the frozen model ($BGE'$) contributed 0.35 to the overall output, while the fine-tuned model ($BGE$) contributed 0.65. All experiments were conducted using g5.12xlarge GPUs.

\section{Results \& Discussion}

Our evaluation is based on the metrics defined in Section \ref{sec:eval_met}. We present our findings in two parts: 1) The enhancement in retrieval performance achieved through the generation of synthetic data and the model fusion approach. 2) We discuss how the REFINE approach not only outperforms other baseline methods but also enables the model to maintain its general capability, thereby mitigating catastrophic forgetting

The key findings from the part 1 experiment, where results were obtained on the TOURISM (private), SQUAD (public) and RAG (public) datasets, are presented in Table \ref{tab:part_1}. Our approach of using synthetic data combined with model fusion boosted the performance of the retriever significantly, leading to improvements of \textbf{5.79\%}, \textbf{6.58\%} and \textbf{0.32\%} in Recall@3 for the TOURISM, SQUAD and RAG datasets, respectively compared to the baseline Vanilla BGE model. The gain in recall did not come at the expense of precision (MAP), which also improved.  The substantial improvement observed on the TOURISM dataset highlights the importance of fine-tuning in specific domain settings. Furthermore, even the standard BGE fine-tuning on synthetic data led to improved metrics across both datasets, except for recall on the RAG dataset. An important observation was that the TOURISM dataset was comprised mostly of structured tabular data. Upon further analysis, we observed that the vanilla BGE failed to correctly identify documents for the queries, especially when the queries involved specific details within tables. For instance, when querying "Which hotel's phone number is xxxx?" where xxxx refers to a value within a table cell, the vanilla BGE model struggled to provide accurate representations, resulting in unreliable cosine similarity. However, with augmented data used for fine-tuning, performance improved notably, particularly in the case of structured data retrieval.

\begin{table*}
  \caption{
    Results on TOURISM, RAG, and SQUAD datasets using the baselines defined in the paper and REFINE (our approach). For all evaluation metrics, higher values indicate better performance.
  }
  \resizebox{12.2cm}{!} {
    \centering
    \setlength{\tabcolsep}{1.5pt} 
    \renewcommand{\arraystretch}{1.1} 
    \begin{tabular}{|l|llll|llll|llll|}
      \hline
      \multicolumn{1}{|c|}{}                                & \multicolumn{4}{c|}{\textbf{TOURISM Dataset}}                                                                                                                        & \multicolumn{4}{c|}{\textbf{RAG Dataset}}                                                                                                                          & \multicolumn{4}{c|}{\textbf{SQUAD}}                                                                                                                     \\ \cline{2-13} 
      \multicolumn{1}{|c|}{\multirow{-1}{*}{\textbf{Model}}} & \textbf{MAP} & \textbf{NDCG} & \textbf{MRR} & \textbf{Recall} & \textbf{MAP} & \textbf{NDCG} & \textbf{MRR} & \textbf{Recall} & \textbf{MAP} & \textbf{NDCG} & \textbf{MRR} & \textbf{Recall} \\ 
      & \textbf{@3} & \textbf{@3} & \textbf{@3} & \textbf{@3} & \textbf{@3} & \textbf{@3} & \textbf{@3} & \textbf{@3} & \textbf{@3} & \textbf{@3} & \textbf{@3} & \textbf{@3} \\ \hline
      Vanilla BGE                                          & 0.781       & 0.800       & 0.789       & 0.884         & 0.863       & 0.858       & 0.904       & 0.937         & 0.763       & 0.789       & 0.763       & 0.866         \\ \hline
      snowflake-arctic-embed-m                             & 0.794       & 0.817       & 0.796       & 0.888         & 0.874       & 0.857       & \textbf{0.921} & 0.916         & 0.708       & 0.728       & 0.707       & 0.786         \\ \hline
      text-embedding-3-large                               & 0.847       & 0.857       & 0.850       & 0.915         & 0.878       & 0.866       & 0.920       & 0.935         & 0.808       & 0.828       & 0.808       & 0.883         \\ \hline
      Fine-tuned BGE                                       & 0.870       & 0.883       & 0.873       & 0.929         & 0.876       & 0.865       & 0.915       & 0.929         & 0.834       & 0.853       & 0.834       & 0.907         \\ \hline
      LM-cocktail (BGE+FT-BGE)                            & 0.862       & 0.881       & 0.869       & 0.936         & 0.870       & 0.861       & 0.906       & 0.934         & 0.829       & 0.852       & 0.829       & 0.920         \\ \hline
      REFINE (ours)                                         & \textbf{0.873} & \textbf{0.885} & \textbf{0.877} & \textbf{0.937} & \textbf{0.881} & \textbf{0.867} & 0.919       & \textbf{0.940} & \textbf{0.846} & \textbf{0.866} & \textbf{0.846} & \textbf{0.923} \\ \hline                  
    \end{tabular}
  }

  \label{tab:part_1}
\end{table*}

\vspace{-1.2cm}

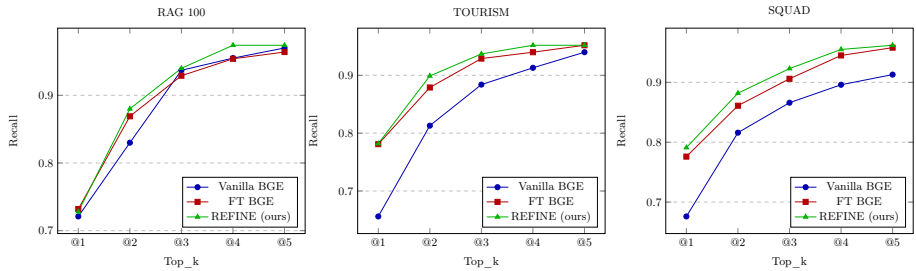
\begin{figure*}[ht]
    \centering
    \begin{tikzpicture}[scale=0.48]
    
    \begin{axis}[
        title={RAG 100},
        xlabel={Top\_k},
        ylabel={Recall},
        xtick=data,
        xticklabels={@1, @2, @3, @4, @5},
        legend pos=south east,
        ymajorgrids=true,
        grid style=dashed,
    ]
    
    \addplot[color=blue!70!black, mark=*] coordinates {
        (1, 0.721)
        (2, 0.830)
        (3, 0.937)
        (4, 0.955)
        (5, 0.970)
    };
    \addlegendentry{Vanilla BGE}

    \addplot[color=red!70!black, mark=square*] coordinates {
        (1, 0.732)
        (2, 0.869)
        (3, 0.929)
        (4, 0.954)
        (5, 0.964)
    };
    \addlegendentry{FT BGE}
    
    \addplot[color=green!70!black, mark=triangle*] coordinates {
        (1, 0.728)
        (2, 0.880)
        (3, 0.940)
        (4, 0.974)
        (5, 0.974)
    };
    \addlegendentry{REFINE (ours)}

    \end{axis}
    
    \end{tikzpicture}%
    \begin{tikzpicture}[scale=0.48]
    \begin{axis}[
        title={TOURISM},
        xlabel={Top\_k},
        ylabel={Recall},
        xtick=data,
        xticklabels={@1, @2, @3, @4, @5},
        legend pos=south east,
        ymajorgrids=true,
        grid style=dashed,
    ]
    
    \addplot[color=blue!70!black, mark=*] coordinates {
        (1, 0.656)
        (2, 0.813)
        (3, 0.884)
        (4, 0.913)
        (5, 0.940)
    };
    \addlegendentry{Vanilla BGE}
    
    \addplot[color=red!70!black, mark=square*] coordinates {
        (1, 0.781)
        (2, 0.879)
        (3, 0.929)
        (4, 0.940)
        (5, 0.952)
    };
    \addlegendentry{FT BGE}
    
    \addplot[color=green!70!black, mark=triangle*] coordinates {
        (1, 0.782)
        (2, 0.899)
        (3, 0.937)
        (4, 0.952)
        (5, 0.952)
    };
    \addlegendentry{REFINE (ours)}

    \end{axis}
    \end{tikzpicture}
\begin{tikzpicture}[scale=0.48]

    \begin{axis}[
        title={SQUAD},
        xlabel={Top\_k},
        ylabel={Recall},
        xtick=data,
        xticklabels={@1, @2, @3, @4, @5},
        legend pos=south east,
        ymajorgrids=true,
        grid style=dashed,
    ]
    
    \addplot[color=blue!70!black, mark=*] coordinates {
        (1, 0.676)
        (2, 0.816)
        (3, 0.866)
        (4, 0.896)
        (5, 0.913)
    };
    \addlegendentry{Vanilla BGE}
    
    \addplot[color=red!70!black, mark=square*] coordinates {
        (1, 0.776)
        (2, 0.861)
        (3, 0.906)
        (4, 0.945)
        (5, 0.958)
    };
    \addlegendentry{FT BGE}
    
    \addplot[color=green!70!black, mark=triangle*] coordinates {
        (1, 0.791)
        (2, 0.882)
        (3, 0.923)
        (4, 0.955)
        (5, 0.962)
    };
    \addlegendentry{REFINE (ours)}

    \end{axis}
    
    \end{tikzpicture}
    \caption{Recall at different top\_k values for vanilla (pretrained) BGE, fine-tuned (FT) BGE, and REFINE.}
    \label{fig:line-chart}
\end{figure*}


\vspace{-0.5cm}
Figure \ref{fig:line-chart} plots the performance of the three models when different values of top\_k were used in the retriever. Recall was measured with $\mathrm{top\_k} = 1, 2, 3, 4, 5$. The figure shows that REFINE outperforms the other baselines across the different top\_k settings on both datasets. A significant improvement can be observed in the TOURISM and SQUAD dataset, even for $\mathrm{top\_k}=1$ w.r.t vanilla model. This demonstrates the superior performance of REFINE when retrieving only one or few relevant documents. Refer Table \ref{sup:recall_table} in Supplementary material for the detailed results. We also want to highlight that in preliminary experiments, we attempted unsupervised fine-tuning on the unlabeled training data. However, this approach led to a degradation in performance. The reason for this is that the objective of pre-training is to reconstruct the text, and the pre-trained model cannot be directly used for similarity computation. Instead, it requires fine-tuning to achieve that capability \cite{xiao2024cpack, FlagOpen2024FlagEmbedding}.

\textbf{Cross-Dataset Generalization.}

We further conducted experiments to evaluate the impact of fine-tuning on the generalizability of a model trained on domain-specific data. While fine-tuning generally improves performance within the targeted domain (in-domain), it often results in a loss of generalization across other domains (out-of-domain), a phenomenon known as catastrophic forgetting \cite{aleixo2023catastrophic}. Our findings, presented in Table \ref{tab:part_2}, show the performance of a model initially trained on the SQUAD dataset and then evaluated on the RAG dataset, serving as an out-of-domain test case (OOD). We also compared our results with the LM-cocktail method, a popular model-merging technique. Our approach improved performance, achieving a recall of \textbf{0.938} compared to baseline methods. Interestingly, even fine-tuned BGE exhibited improved rather than degraded performance. We want to highlight that our primary focus was on enhancing the retriever for the specific dataset, we aimed to examine the effects of catastrophic forgetting and acknowledge the need for further exploration in future studies. Therefore, we limited our analysis to the comparison between the SQuAD and RAG datasets, rather than examining all possible combinations. Additionally, when we reversed the training and test datasets (using RAG as the training data and SQUAD as the test data), a similar trend was observed. Fine-tuned BGE, LM-cocktail showed performance improvements, with REFINE outperforming others achieving a recall of \textbf{0.896}, highlighting the generalizability of our approach.

\begin{table*}
  \caption{
    The results shown represent the model's performance when it was trained on a specific dataset and evaluated on an out-of-domain (OOD) dataset. The row with a "-" indicates the metrics for the vanilla (pretrained) BGE model, which did not undergo fine-tuning.
  }
  \centering
\begin{tabular}{|c|c|c|cccc|}
\hline
{ }                                         & { }                                        & { }                                      & { }                               & { }                                & { }                               & { }                                  \\
\multirow{-1}{*}{{ \textbf{Train Dataset}}} & \multirow{-1}{*}{{ \textbf{OOD Dataset}}} & \multirow{-1}{*}{{ \textbf{Model}}} & \multirow{-2}{*}{{ \textbf{MAP}}} & \multirow{-2}{*}{{ \textbf{NDCG}}} & \multirow{-2}{*}{{ \textbf{MRR}}} & \multirow{-2}{*}{{ \textbf{Recall}}} \\

\multirow{-2}{*}{{ \textbf{}}} & \multirow{-2}{*}{{ \textbf{}}} & \multirow{-2}{*}{{ \textbf{}}} & \multirow{-2}{*}{{ \textbf{@3}}} & \multirow{-2}{*}{{ \textbf{@3}}} & \multirow{-2}{*}{{ \textbf{@3}}} & \multirow{-2}{*}{{ \textbf{@3}}} \\ 

\hline
{ -} & { } & { Vanilla BGE} & { 0.863} & { 0.858} & { 0.904} & { 0.937} \\ \cline{1-1} \cline{3-7} 
{ }  & { } & { FT BGE} & { 0.879} & { 0.866} & { 0.916} & { 0.932} \\ \cline{3-7} 

{ }  & { } & {LM-cocktail} & { 0.878} & { \textbf{0.866}} & { \textbf{0.917}} & { 0.934} \\ \cline{3-7} 
\multirow{-3}{*}{{ SQUAD}} & \multirow{-4}{*}{{ RAG}} & { REFINE (ours)} & { \textbf{0.881}} & { 0.865} & { 0.915} & {\textbf{0.938}} \\ \hline

- & & Vanilla BGE & 0.708 & 0.728 & 0.707 & 0.786                                                   \\ \cline{1-1} \cline{3-7} 
                                                                &      & FT BGE          & 0.743    & 0.771     & 0.743         & 0.852      \\ \cline{3-7} 

                                                                &      & LM-cocktail          & 0.769    & 0.794     & 0.769         & 0.866      \\ \cline{3-7} 
\multirow{-3}{*}{RAG}                                       & \multirow{-4}{*}{SQUAD}                                    & REFINE (ours)                                                     & \textbf{0.801}                                        & \textbf{0.825}                                         & \textbf{0.801}                                        & \textbf{0.896}                                           \\ \hline
\end{tabular}

  \label{tab:part_2}
\end{table*}


\vspace{-0.5cm}

\section{Conclusions and Future Work}

In this work, we introduced a novel approach to improve retrieval models used in RAG. First, we generated synthetic contrastive data to fine-tune the embedding models.  Then, we introduced a model fusion approach, which combines the pretrained (vanilla) model with the newly trained model during the training phase. This fusion method led to significant performance gains across various metrics and improved the model's ability to generalize to unseen data. Although this study focused on scarce data to simulate real-life scenarios, we believe our approach can be extended to larger datasets, potentially achieving even greater performance improvements. In the future, we plan to apply our technique to more (larger) datasets and further investigate catastrophic forgetting in retrieval embedding models. Additionally, we aim to explore multi-hop retrieval fine-tuning, which is a challenging task.

\bibliographystyle{splncs04}
\bibliography{custom}

\appendix


\section{Detailed Results}
The detailed recall results at different top\_k values for both datasets are presented in Table \ref{sup:recall_table}. Our approach significantly outperforms all the baselines except in two cases. For the TOURISM dataset, LM-cocktail performed slightly better than REFINE for $\mathrm{top\_k} = 5$, while \texttt{snowflake-arctic-embed-l} performed better for $\mathrm{top\_k} = 1$ in case of RAG. Notably, on the TOURISM \& SQUAD dataset, OpenAI's \texttt{text-embedding-3-large} embedding model demonstrated significantly better performance than other embedding models used in a vanilla (without fine-tuning) setting.

\begin{table*}[]
\caption{
    Recall at different top\_k values for the baselines defined in the paper and REFINE (our approach).
  }
\resizebox{12.2cm}{!} {
\begin{tabular}{|l |l |lllll |}
\hline
\multicolumn{1}{|c|}{}                                   & \multicolumn{1}{c|}{}                                 & \multicolumn{5}{c|}{{\textbf{Recall}}}                                        \\ \cline{3-7} 
\multicolumn{1}{|c|}{\multirow{-2}{*}{{ \textbf{Dataset}}}} & \multicolumn{1}{c|}{\multirow{-2}{*}{{ \textbf{Model}}}} & { \textbf{@1}}    & { \textbf{@2}}    & { \textbf{@3}}    & { \textbf{@4}}    & { \textbf{@5}}    \\ \hline
{ }                                                         & { Vanilla BGE}                                                                   & { 0.676}          & { 0.816}          & { 0.866}          & { 0.896}          & { 0.913}          \\ \cline{2-7} 
{ }                                                         & { snowflake-arctic-embed-l}                                                      & { 0.643}          & { 0.743}          & { 0.786}          & { 0.823}            & { 0.850}            \\ \cline{2-7} 
{ }                                                         & { text-embedding-3-large}                                                        & { 0.743}          & { 0.856}          & { 0.883}          & { 0.903}          & { 0.916}          \\ \cline{2-7} 
{ }                                                         & { Fine-tuned BGE}                                                                & { 0.776}          & { 0.861}          & { 0.906}          & { 0.945}           & { 0.958}          \\ \cline{2-7} 
{ }                                                         & { LM-cocktail}                                                                   & { 0.781}          & { 0.862}          & { 0.908}          & { 0.946}          & { 0.959} \\ \cline{2-7} 
\multirow{-6}{*}{{SQUAD}}                              & { REFINE (ours)}                                                                 & { \textbf{0.791}} & { \textbf{0.882}} & { \textbf{0.923}} & { \textbf{0.955}} & { \textbf{0.962}}          \\ \hline
{ }                                                         & { Vanilla BGE}                                                                   & { 0.656}          & { 0.813}          & { 0.884}          & { 0.913}          & { 0.940}          \\ \cline{2-7} 
{ }                                                         & { snowflake-arctic-embed-l}                                                      & { 0.665}          & { 0.840}          & { 0.888}          & { 0.9}            & { 0.9}            \\ \cline{2-7} 
{ }                                                         & { text-embedding-3-large}                                                        & { 0.738}          & { 0.895}          & { 0.915}          & { 0.936}          & { 0.954}          \\ \cline{2-7} 
{ }                                                         & { Fine-tuned BGE}                                                                & { 0.781}          & { 0.879}          & { 0.929}          & { 0.94}           & { 0.952}          \\ \cline{2-7} 
{ }                                                         & { LM-cocktail}                                                                   & { 0.762}          & { 0.872}          & { 0.936}          & { 0.946}          & { \textbf{0.955}} \\ \cline{2-7} 
\multirow{-6}{*}{{TOURISM}}                              & { REFINE (ours)}                                                                 & { \textbf{0.782}} & { \textbf{0.899}} & { \textbf{0.937}} & { \textbf{0.952}} & { 0.952}          \\ \hline
{ }                                                         & { Vanilla BGE}                                                                   & { 0.721}          & { 0.83}           & { 0.937}          & { 0.955}          & { 0.97}           \\ \cline{2-7} 
{ }                                                         & { snowflake-arctic-embed-l}                                                      & { \textbf{0.739}} & { 0.865}          & { 0.916}          & { 0.954}          & { 0.954}          \\ \cline{2-7} 
{ }                                                         & { text-embedding-3-large}                                                        & { 0.734}          & { 0.865}          & { 0.935}          & { 0.959}          & { 0.964}          \\ \cline{2-7} 
{ }                                                         & { Fine-tuned BGE}                                                                & { 0.732}          & { 0.869}          & { 0.929}          & { 0.954}          & { 0.964}          \\ \cline{2-7} 
{ }                                                         & { LM-cocktail}                                                                   & { 0.720}          & { 0.859}          & { 0.934}          & { 0.962}          & { 0.963}          \\ \cline{2-7} 
\multirow{-6}{*}{{ RAG 100}}                                & { REFINE (ours)}                                                                 & { 0.728}          & { \textbf{0.88}}  & { \textbf{0.94}}  & { \textbf{0.974}} & { \textbf{0.974}} \\ \hline

\end{tabular}
}
 
  \label{sup:recall_table}
\end{table*}

\newpage

\end{document}